\documentclass{article}

\usepackage[final]{neurips_2025}
\usepackage{booktabs}
\usepackage{tabularx}
\usepackage{ragged2e}  
\usepackage{amsmath}
\usepackage{enumitem}
\usepackage{amssymb} 
\usepackage{amsthm}
\usepackage{graphicx}
\usepackage{microtype}
\usepackage{adjustbox}
\usepackage{multicol}
\usepackage{multirow}
\usepackage{colortbl} 
\usepackage{pifont}    % For checkmarks and crosses
\usepackage{tcolorbox}
\usepackage{wrapfig}
\usepackage[table]{xcolor}

\usepackage[utf8]{inputenc} % allow utf-8 input
\usepackage[T1]{fontenc}    % use 8-bit T1 fonts
\usepackage{hyperref}       % hyperlinks
\usepackage{url}            % simple URL typesetting
\usepackage{amsfonts}       % blackboard math symbols
\usepackage{nicefrac}       % compact symbols for 1/2, etc.
\usepackage{xcolor}         % colors

\title{Event Fields: Learning Latent Event Structure for Waveform Foundation Models}

\author{
Li Na$^1$ ,Yuanyun Zhang$^1$, Shi Li$^2$\\
$^1$ University of Science and Technology of China\\
$^2$ Columbia University
}
\newcolumntype{Y}{>{\RaggedRight\arraybackslash}X}

\begin{document}

\maketitle

\begin{abstract}
We propose a new class of waveform foundation models that departs from conventional sequence-based representations by modeling physiological time series as realizations of latent event processes. Rather than treating signals as collections of local tokens or patches, our approach assumes that clinically meaningful structure arises from temporally extended, interacting events whose boundaries and dynamics are not directly observed. To capture this structure, we introduce a self-supervised learning framework that enforces consistency across stochastic segmentations and time–frequency projections of the same waveform, encouraging representations that are invariant to signal-level perturbations while preserving event-level organization. The resulting model combines a segmentation-aware encoder with a latent interaction operator that captures dependencies among inferred events, and naturally extends to multimodal settings by aligning modalities through shared event representations. Across a range of physiological benchmarks, including arrhythmia classification, hemodynamic prediction, and waveform retrieval, the proposed method improves performance, robustness, and label efficiency relative to strong sequence-based baselines. These results suggest that shifting from signal-centric to event-centric representations provides a more appropriate inductive bias for modeling physiological dynamics and offers a complementary path to scaling foundation models in healthcare.

\end{abstract}

\section{Introduction}

Recent progress in healthcare foundation models has largely adhered to a now-familiar paradigm: increasing model scale, pretraining on large heterogeneous corpora, and transferring learned representations to downstream clinical tasks \cite{thieme2023foundation, vaid2023foundational, he2024foundation, burkhart2025foundation}. 
This recipe has been successfully instantiated across modalities, including structured EHR, imaging, and physiological waveforms, with a growing body of work emphasizing the importance of scale and data diversity \cite{guo2025foundation, liang2024foundation, awais2025foundation, burger2025foundation, thakur2024foundation}. 
For continuous biosignals such as electrocardiography (ECG), photoplethysmography (PPG), and electromyography (EMG), prevailing approaches adopt sequence modeling strategies, typically relying on masked reconstruction or contrastive objectives inspired by language and vision pretraining \cite{devlin2019bert, he2022masked}. 
These models further inherit architectural design principles—such as residual connections and hierarchical feature aggregation—from advances in deep convolutional and transformer-based systems \cite{he2015deepresiduallearningimage, he2017multi, he2019bag}. 
Despite strong empirical performance, such formulations fundamentally treat waveforms as collections of tokens or patches, implicitly assuming that predictive structure is encoded directly in local signal segments.

This assumption is misaligned with the nature of physiological reasoning. Clinical interpretation of waveforms rarely proceeds at the level of raw amplitudes or short-time windows; instead, it is mediated through higher-level abstractions such as beats, cycles, bursts, and transitions between dynamical regimes. 
These abstractions correspond to temporally extended, structured phenomena that are only indirectly observable in the signal. 
For instance, cardiac function is understood through the organization of successive beats, vascular compliance through pulse transit dynamics, and neuromuscular activity through coordinated activation patterns. 
Crucially, these phenomena are inherently relational and temporal: their meaning arises not from isolated occurrences, but from their ordering, durations, and interactions. 
Standard sequence models must implicitly recover such structure from low-level supervision, often conflating meaningful variability with nuisance signal fluctuations.

In this work, we introduce a new class of waveform foundation models grounded in a \emph{spectro-temporal event field} formulation. 
Rather than modeling the waveform \(x(t)\) directly in the time domain, we posit that physiological signals are generated by a continuous superposition of latent event fields evolving jointly in time and frequency. 
Formally, we represent the signal as
\[
x(t) = \int_{\Omega} \int_{0}^{T} \phi(z, \tau) \, g_\theta(t - \tau, z)\, d\tau \, dz,
\]
where \(z \in \Omega\) indexes a latent event space (capturing attributes such as rhythm class, activation type, or modality-specific structure), \(\phi(z, \tau)\) denotes an event intensity field over latent states and time, and \(g_\theta\) is a learned emission kernel that maps latent events into waveform contributions. 
This formulation generalizes discrete event models by allowing events to overlap, interact, and evolve continuously, thereby capturing both transient phenomena and sustained dynamical regimes.

The central modeling challenge is that the latent event field \(\phi(z, t)\) is unobserved. 
To address this, we propose a self-supervised learning objective based on \emph{spectro-temporal consistency under stochastic projections}. 
Specifically, we generate multiple views of the same waveform via randomized time-frequency projections (e.g., sub-band filtering, time warping, or phase perturbations), and train an encoder \(h_\theta\) to produce representations that are invariant to these projections while preserving the underlying event field:
\[
\mathcal{L}_{\mathrm{field}} =
\mathbb{E}_{x,\, \mathcal{P}_1, \mathcal{P}_2}
\left[
- \log \frac{\exp\big(\langle h_\theta(\mathcal{P}_1(x)),\, h_\theta(\mathcal{P}_2(x)) \rangle / \tau \big)}
{\sum_{x'} \exp\big(\langle h_\theta(\mathcal{P}_1(x)),\, h_\theta(x') \rangle / \tau \big)}
\right].
\]
Unlike conventional augmentations that operate in the signal domain, these projections selectively distort time-frequency structure, forcing the model to recover representations that correspond to invariant latent event fields rather than superficial waveform characteristics.

To further capture the structure of physiological dynamics, we introduce a \emph{continuous event operator} that models interactions within the latent field. 
This operator acts as a neural integral transform over \(\phi(z, t)\), enabling the model to represent dependencies such as periodic coupling, cross-frequency interactions, and modality alignment. 
Concretely, the encoder produces a function-valued representation over latent states, and interactions are modeled via kernelized operators:
\[
\mathcal{K}_\theta[\phi](z, t) = \int_{\Omega} \int_{0}^{T} k_\theta(z, z', t - t') \, \phi(z', t') \, dt' \, dz'.
\]
This construction unifies ideas from neural operators and attention mechanisms, while providing an explicit inductive bias toward structured temporal interactions. 
In multimodal settings, different waveforms correspond to distinct projections of a shared latent field, allowing alignment to emerge naturally through shared event representations rather than explicit fusion modules.

Our formulation induces a fundamentally different notion of invariance. 
Instead of enforcing robustness to local perturbations in the signal domain, the model is encouraged to identify stable structures in the latent event field that persist across time-frequency distortions and modality-specific emissions. 
This yields representations that are sensitive to clinically meaningful changes—such as alterations in rhythm or coupling—while discarding nuisance variability such as noise, baseline drift, or sensor-specific artifacts.

Beyond representation quality, the proposed framework introduces a new scaling perspective. 
A single waveform admits a combinatorially large family of time-frequency projections, each providing a distinct but consistent view of the same latent field. 
This effectively amplifies the available supervision signal without requiring additional data, complementing traditional scaling strategies based on dataset size and model capacity \cite{he2019bag, he2017multi, he2015deepresiduallearningimage}. 
Moreover, the continuous nature of the latent field allows the model to interpolate across temporal and spectral resolutions, facilitating transfer across tasks with varying sampling rates and modalities.

The contributions of this work are threefold. 
First, we propose a spectro-temporal event field formulation for physiological waveforms, shifting the modeling target from raw signals to continuous latent processes. 
Second, we introduce a self-supervised objective based on stochastic time-frequency projections that enforces invariance at the level of latent event structure. 
Third, we develop a neural operator-based architecture that captures interactions within the latent field, enabling principled modeling of temporal dependencies and multimodal alignment. 
Together, these components define a new class of waveform foundation models that move beyond tokenized sequence representations and instead learn structured, continuous abstractions of physiological dynamics.

\section{Related Works}

Self-supervised learning (SSL) has become the central paradigm for representation learning in healthcare, enabling the development of foundation models across diverse modalities such as clinical narratives, structured EHRs, medical imaging, and physiological waveforms \cite{lee2024can, chou2025serialized, zhang2025collection, huiliang2025clio, ran2025structured, lee2025towards, lin2025case, lowelatent, zhang2025chronoformer}. 
These approaches share a unifying principle: large-scale pretraining on unlabeled data yields transferable representations that can be adapted to a wide range of downstream clinical tasks. 
This trajectory mirrors advances in natural language processing and computer vision, where masked reconstruction and generative pretraining objectives \cite{devlin2019bert, he2022masked, lee2025himae}, combined with increasingly deep and expressive architectures \cite{he2017multi, he2015deepresiduallearningimage, he2019bag}, have produced general-purpose representations with strong transferability. 
However, despite their empirical success, these methods typically treat time series as generic sequences of tokens or patches, thereby neglecting the structured and event-driven nature of physiological signals that underlies our formulation.

A substantial body of work has focused on \emph{temporal and structured modeling} of longitudinal clinical data. 
Sequence architectures such as Chronoformer and related models explicitly encode temporal dependencies and hierarchical organization within patient trajectories \cite{ran2025structured, zhang2025chronoformer, zhang2025collection}. 
In the context of high-frequency biosignals, classical signal processing priors—including time-frequency decompositions and multi-resolution analysis—have been incorporated to capture oscillatory and multi-scale dynamics \cite{oppenheim1999discrete, daubechies1992ten}. 
Recent large-scale models for wearable and clinical time series extend these ideas by combining deep architectures with domain-specific priors \cite{yang2023biot, abbaspourazad2023large, abbaspourazad2024wearable, lee2025himae, lee2025foundation}. 
While these approaches improve inductive bias by embedding structure into the model, they remain fundamentally signal-centric, operating directly on observed waveforms without explicitly modeling the latent processes that generate them. 
In contrast, our framework elevates latent event dynamics to the primary object of representation, enforcing consistency across alternative segmentations rather than relying on fixed temporal discretizations.

Another prominent direction is \emph{generative and predictive modeling}, in which models are trained to approximate the distribution of observed clinical data. 
Masked modeling reconstructs missing tokens, while autoregressive approaches capture sequential dependencies in patient records \cite{brown2020language}. 
These paradigms have been extensively adapted to EHR data, supporting tasks such as diagnosis prediction, clinical event modeling, and synthetic data generation \cite{steinberg2021language, rasmy2021med, wornow2023shaky, fallahpour2024ehrmamba, lee2024emergency, ono2024text, steinberg2024motor, lee2025clinical}. 
In physiological time series, analogous objectives are applied in both time and frequency domains, often assuming that accurate reconstruction implies meaningful representation learning. 
Our approach departs from this assumption by positing that clinically relevant structure is not captured by signal fidelity alone, but rather by the organization and interaction of latent events. 
Accordingly, we replace reconstruction-based supervision with a consistency objective that operates over stochastic segmentations and projections.

Complementing generative approaches, \emph{contrastive and alignment-based methods} have been widely adopted to learn invariant representations through similarity constraints \cite{bertram2024contrastivelearningpreferencescontextual, tian_2019_contrastic_distillation}. 
These methods align augmented views of the same sample while separating representations of different samples, and have been extended to multimodal clinical settings via shared embedding spaces and cross-attention mechanisms \cite{hou2019cross, chen2021crossvit, huang2019ccnet, wornow2024context, odgaard2024core, shmatko2025learning}. 
This paradigm is closely connected to progress in vision and multimodal learning \cite{radford2021learning, caron2021emerging, zhou2021ibot, saharia2022photorealistic, rombach2021highresolution, oquab2023dinov2, Ranftl2022, kirillov2023segment, liu2024visual}, where invariance to augmentations yields semantically meaningful embeddings. 
Our formulation can be interpreted as a shift from view-based alignment to \emph{structure-based consistency}: rather than aligning augmented signals, we enforce agreement across stochastic segmentations and time-frequency projections, thereby encouraging the discovery of intrinsic event structure.

At the architectural level, advances in \emph{scaling and efficiency} have played a crucial role in enabling modern foundation models. 
Transformer-based architectures and their variants dominate large-scale representation learning \cite{dosovitskiy2021an, liu2021swin}, supported by efficient attention mechanisms \cite{dao2023flashattention2} and hybrid designs tailored for high-dimensional and long-context data \cite{choy20194d, wu2023e2enet, lai2024e3d, liu2024octcube, shaker2024unetr++, xing2024segmamba, lee2025modern}. 
These developments have made it feasible to model long, high-resolution physiological sequences, including continuous monitoring data. 
Nevertheless, increasing model capacity does not by itself resolve the mismatch between sequence-based representations and the latent event-driven structure of biosignals. 
Our work is complementary to these advances: it introduces a new learning objective and representation space that can be instantiated within scalable architectures while fundamentally altering the abstraction at which learning occurs.

Related ideas have also emerged in \emph{biological sequence modeling}, where SSL is applied to genomic and molecular data \cite{ji2021dnabert, le2021transformer, ma2025hybridna, larey2026jepa, liu2024genbench, wu2025generator, lin2025genos, kolo2024meds}. 
These approaches demonstrate the importance of incorporating domain-specific structure—such as motifs, folding constraints, or evolutionary patterns—into representation learning. 
Our work extends this perspective to physiological waveforms, where the relevant structure arises from latent, temporally evolving event processes rather than discrete symbolic sequences.

Across these lines of work, a common assumption is that representations should encode all relevant information about patient state directly from observed data, whether through reconstruction, prediction, or alignment. 
In contrast, we model physiological signals as realizations of an underlying latent event field and emphasize invariance across alternative segmentations and projections of the same process. 
This shift from signal-centric to event-centric learning introduces a new inductive bias that prioritizes structural consistency over observational fidelity, suggesting a different pathway for improving representation learning in healthcare time series.

Finally, evaluation remains a central challenge for healthcare foundation models, particularly in assessing generalization, robustness, and clinical utility \cite{donabedian2005evaluating, mcdermott2025meds, gray2020medical, bedi2025medhelm, zhao2023survey, singhal2023large}. 
As models become increasingly abstract and rely on latent representations, developing benchmarks that capture clinically meaningful structure will be essential for validating emerging paradigms such as the one proposed here.

\section{Methods}

\begin{figure*}[t!]
    \centering
    \includegraphics[width=\linewidth]{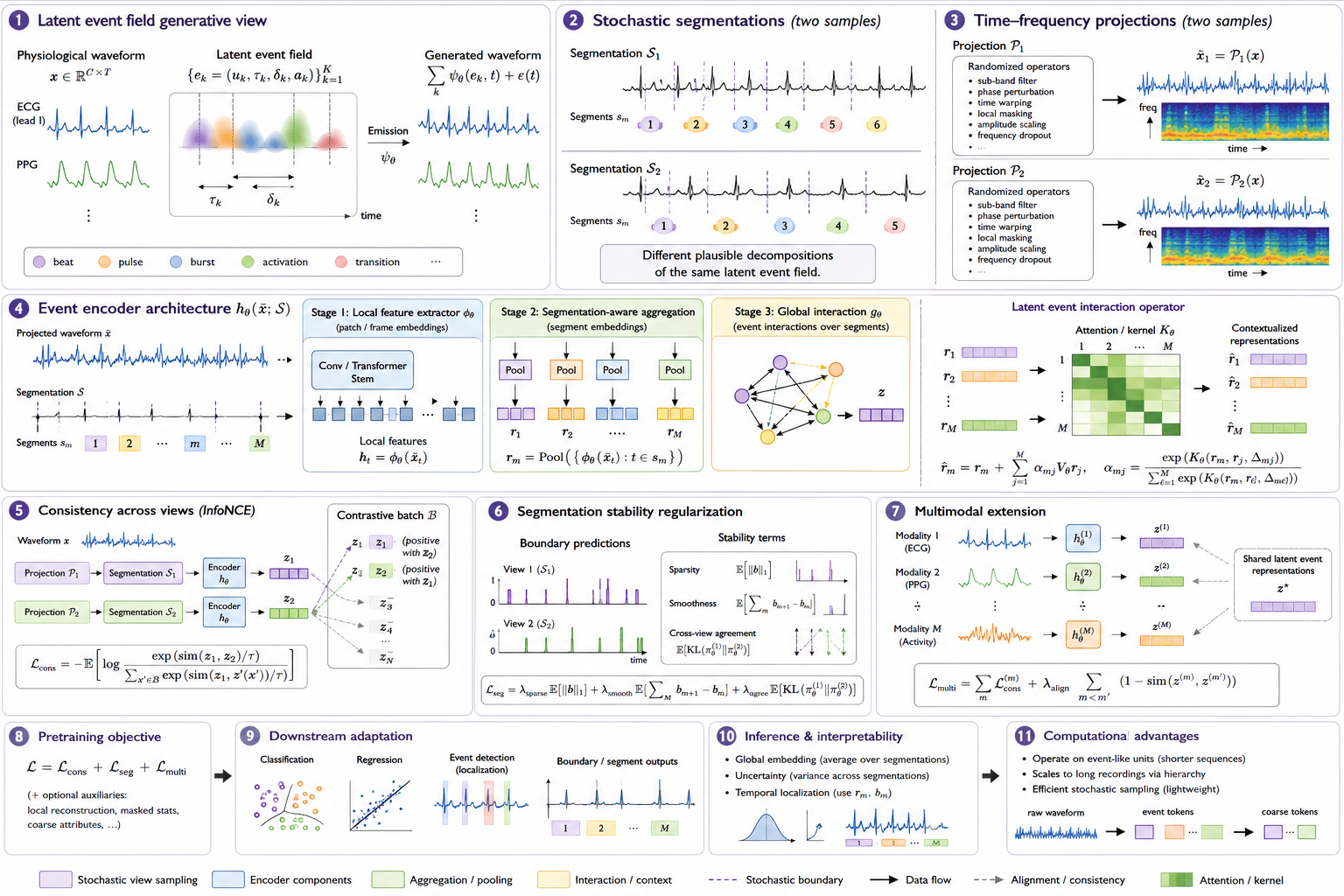}
    \caption{O\textbf{verview of the proposed event-field waveform foundation model. }
(1) \textbf{Latent event field generative view:} physiological waveforms are modeled as superpositions of latent events with variable durations and emissions. 
(2) \textbf{Stochastic segmentations:} multiple plausible decompositions of the same signal are sampled, reflecting ambiguity in event boundaries. 
(3) \textbf{Time--frequency projections:} randomized operators generate complementary views that preserve underlying physiological structure while perturbing signal-level details. 
(4) \textbf{Event encoder architecture:} projected signals are processed through a hierarchical encoder consisting of local feature extraction, segmentation-aware aggregation, and a global interaction module with an event interaction operator. 
(5) \textbf{Consistency objective:} representations from different segmentations and projections are aligned via a contrastive loss. 
(6) \textbf{Segmentation regularization:} boundary predictions are constrained by sparsity, smoothness, and cross-view agreement. 
(7) \textbf{Multimodal extension:} multiple modalities are treated as emissions of a shared latent event field and aligned in a common representation space. 
(8) \textbf{Pretraining objective:} the model is trained with a combination of consistency, segmentation, and multimodal alignment losses. 
(9) \textbf{Downstream adaptation:} learned representations support classification, regression, and event localization tasks. 
(10) \textbf{Inference and interpretability:} the model produces stable global embeddings, uncertainty estimates via segmentation variability, and temporally localized event representations. 
(11) \textbf{Computational advantages:} event-centric representations enable efficient processing of long waveforms through hierarchical abstraction. 
}
    \label{fig:placeholder}
\end{figure*}

\paragraph{Overview.}
We propose a waveform foundation model that learns representations from physiological signals by treating them as realizations of an underlying latent event field. 
The core hypothesis is that clinically meaningful structure is not most naturally expressed at the level of raw samples, but at the level of temporally extended events such as beats, bursts, activations, transitions, and coupled responses. 
Accordingly, our model is designed to infer latent event structure from unlabeled waveforms and to produce representations that remain stable across alternative stochastic segmentations and time-frequency projections of the same signal. 
This section describes the latent generative perspective, the event encoder, the stochastic projection mechanism, the self-supervised learning objective, and the multimodal extension used to integrate heterogeneous physiological channels.

\paragraph{Notation and data representation.}
Let \(x \in \mathbb{R}^{C \times T}\) denote a multichannel waveform sampled over \(T\) time steps with \(C\) channels. 
In the unimodal case, \(C=1\); in the multimodal case, each channel may correspond to a different sensing modality or lead configuration. 
We assume access to a dataset \(\mathcal{D} = \{x_i\}_{i=1}^N\) of unlabeled physiological recordings. 
The goal is to learn an encoder \(h_\theta\) that maps each waveform to a compact representation \(z \in \mathbb{R}^d\) suitable for downstream prediction, retrieval, and adaptation. 
Unlike standard sequence models, our encoder is trained not to reconstruct masked samples directly, but to infer representations of latent event organization that are invariant to non-semantic changes in segmentation and projection.

\paragraph{Latent event field model.}
We model each waveform as a continuous superposition of latent events. 
Let \(e_k = (u_k, \tau_k, \delta_k, a_k)\) denote the \(k\)-th event, where \(u_k \in \mathbb{R}^{d_e}\) is a latent event state, \(\tau_k\) is the event onset time, \(\delta_k\) is its duration, and \(a_k\) captures event amplitude or interaction parameters. 
The observed waveform is generated by a learned emission function \(\psi_\theta\) that maps latent events into signal space:
\[
x(t) = \sum_{k=1}^{K} \psi_\theta(e_k, t) + \varepsilon(t),
\]
where \(\varepsilon(t)\) denotes residual noise and unmodeled variation. 
This representation is intentionally flexible: events may overlap, vary in duration, and interact through shared latent factors. 
In contrast to discrete tokenization, which imposes a rigid segmentation a priori, the event field formulation leaves temporal abstraction to be learned from data. 
For multichannel signals, \(\psi_\theta\) may emit channel-specific contributions, enabling shared latent events to generate distinct observed patterns across leads or modalities.

\paragraph{Stochastic event segmentations.}
Since latent events are unobserved, we induce them through stochastic segmentations of each waveform. 
Given a recording \(x\), we sample a segmentation operator \(\mathcal{S}\) that partitions the time axis into \(M\) contiguous segments \(\{s_m\}_{m=1}^{M}\). 
A segmentation may be produced by a stochastic boundary process, a learned proposal network, or a hybrid of signal-dependent and random boundary sampling. 
Each segmentation induces a set of segment embeddings \(\{r_m\}_{m=1}^M\) obtained by summarizing local waveform content within each segment. 
The purpose of stochastic segmentation is not to define ground-truth events, but to expose the model to multiple plausible decompositions of the same underlying process. 
If two segmentations correspond to the same latent event organization, then the learned representation should be consistent across them. 
This principle forms the basis of our learning objective.

\paragraph{Time-frequency projections.}
To provide complementary views of the waveform, we apply randomized projections in both time and frequency \cite{fu2025frequency, kara2024freqmae, liu2023frequency}. 
For each recording \(x\), we sample projection operators \(\mathcal{P}\) from a family of transformations that preserve underlying physiological identity while perturbing nuisance characteristics. 
These projections may include sub-band filtering, phase perturbation, time warping, local masking, amplitude scaling, frequency dropout, or combinations thereof. 
We write the projected view as
\[
\tilde{x} = \mathcal{P}(x).
\]
Unlike simple augmentations intended only to increase invariance, the purpose of these projections is to expose the model to distinct views of the same latent event field. 
The model must therefore explain which aspects of the signal persist under projection and which aspects are incidental to the observed waveform geometry. 
This encourages representations that capture event-level structure rather than sensor-specific artifacts or local fluctuations.

\paragraph{Event encoder architecture.}
The encoder \(h_\theta\) maps a projected waveform \(\tilde{x}\) and a segmentation \(\mathcal{S}\) to a representation \(z\). 
We implement \(h_\theta\) as a hierarchical event encoder with three stages. 
First, a local feature extractor \(\phi_\theta\) computes patch- or frame-level embeddings from the raw waveform. 
This component may be realized using a convolutional front-end, a transformer stem, or a hybrid design depending on the target sampling rate and context length. 
Second, a segmentation-aware aggregation module pools local features into segment embeddings. 
Given segment \(s_m\), we compute
\[
r_m = \mathrm{Pool}\big(\{\phi_\theta(\tilde{x}_{t}) : t \in s_m\}\big),
\]
where \(\mathrm{Pool}\) may be mean pooling, attention pooling, or a learned set function. 
Third, a global interaction module models dependencies among segment embeddings and produces the final waveform representation:
\[
z = h_\theta(\tilde{x}; \mathcal{S}) = g_\theta(r_1, \ldots, r_M).
\]
The interaction module \(g_\theta\) may be implemented with self-attention, graph message passing, or an operator-style kernelized layer. 
In all cases, its role is to infer how local events compose into a coherent physiological process.

\paragraph{Latent event interactions.}
A key aspect of physiological waveforms is that events are not independent. 
Beat morphology depends on preceding activity, respiration modulates pulse amplitude, muscular activation interacts with movement, and sensor dynamics can introduce temporal coupling across channels. 
To model these effects, we introduce a latent interaction operator over segment embeddings. 
Let \(R = \{r_m\}_{m=1}^M\). 
We define an interaction kernel \(K_\theta\) and compute context-aware event representations
\[
\hat{r}_m = r_m + \sum_{j=1}^{M} \alpha_{mj} V_\theta r_j,
\qquad
\alpha_{mj} = \frac{\exp\left(K_\theta(r_m, r_j, \Delta_{mj})\right)}{\sum_{\ell=1}^{M} \exp\left(K_\theta(r_m, r_\ell, \Delta_{m\ell})\right)},
\]
where \(\Delta_{mj}\) denotes the temporal distance between segments \(m\) and \(j\). 
This formulation allows the model to capture local dependencies, long-range couplings, and asymmetric temporal relations. 
The final representation \(z\) is obtained by aggregating \(\{\hat{r}_m\}\) through a learned readout function. 
In practice, this architecture behaves as an event-centric analogue of a transformer, but with the crucial distinction that its tokens are not fixed windows of signal; they are stochastic latent events.

\paragraph{Consistency across segmentations.}
Our learning signal is based on the principle that different stochastic segmentations of the same waveform should induce compatible representations. 
Given a waveform \(x\), we sample two segmentations \(\mathcal{S}_1\) and \(\mathcal{S}_2\), and two projections \(\mathcal{P}_1\) and \(\mathcal{P}_2\), yielding views \(\tilde{x}_1 = \mathcal{P}_1(x)\) and \(\tilde{x}_2 = \mathcal{P}_2(x)\). 
The corresponding representations are
\[
z_1 = h_\theta(\tilde{x}_1; \mathcal{S}_1), \qquad z_2 = h_\theta(\tilde{x}_2; \mathcal{S}_2).
\]
We then encourage \(z_1\) and \(z_2\) to be close when they originate from the same underlying recording and far from representations of other recordings in the minibatch. 
This is implemented with an InfoNCE-style objective:
\[
\mathcal{L}_{\mathrm{cons}} = 
-\mathbb{E}\left[
\log
\frac{
\exp\left(\mathrm{sim}(z_1, z_2)/\tau\right)
}{
\sum_{x' \in \mathcal{B}}
\exp\left(\mathrm{sim}(z_1, z'(x'))/\tau\right)
}
\right],
\]
where \(\mathrm{sim}(\cdot,\cdot)\) is cosine similarity or dot product, \(\tau\) is a temperature parameter, \(\mathcal{B}\) is the minibatch, and \(z'(x')\) denotes a representation from a different sample. 
This loss penalizes representations that depend heavily on arbitrary segmentation choices, thereby biasing the model toward latent structure that is stable under re-partitioning. 
In effect, the model is rewarded when two different decompositions of the same physiological process lead to the same semantic interpretation.

\paragraph{Segmentation stability regularization.}
To prevent degenerate solutions in which segmentation is ignored entirely, we introduce an auxiliary stability term that encourages event boundaries to be informative and reproducible. 
Let \(b_m \in \{0,1\}\) denote a predicted boundary variable at time index \(m\), and let \(\pi_\theta(b \mid x)\) be the boundary distribution predicted by a segmentation head. 
We encourage boundary sparsity, temporal smoothness, and cross-view consistency:
\[
\mathcal{L}_{\mathrm{seg}} =
\lambda_{\mathrm{sparse}} \, \mathbb{E}\!\left[\|b\|_1\right]
+
\lambda_{\mathrm{smooth}} \, \mathbb{E}\!\left[\sum_m |b_{m+1} - b_m|\right]
+
\lambda_{\mathrm{agree}} \, \mathbb{E}\!\left[\mathrm{KL}\big(\pi_\theta(\cdot \mid x, \mathcal{S}_1) \,\|\, \pi_\theta(\cdot \mid x, \mathcal{S}_2)\big)\right].
\]
The first term discourages pathological over-segmentation, the second encourages coherent boundaries, and the third aligns predicted boundaries across stochastic views of the same signal. 
This auxiliary loss is not intended to recover annotated events, which are unavailable, but to shape the model’s inductive bias toward meaningful latent partitions. 
In practice, this stabilizes training and improves interpretability of the learned event structure.

\paragraph{Contrastive negatives and batch construction.}
Negative examples are drawn from other waveforms in the minibatch and, when available, from temporally distant segments of the same subject to increase difficulty. 
For longitudinal recordings, we optionally exclude negatives that are highly correlated due to subject identity or nearby acquisition time, since these may correspond to semantically similar physiological states. 
The composition of positives and negatives can be adjusted depending on the downstream target. 
For instance, in arrhythmia modeling, we may prefer harder negatives that share patient identity but differ in rhythm state; for fatigue or motion tasks, we may instead emphasize cross-subject negatives to avoid overfitting to sensor-specific idiosyncrasies. 
The resulting objective remains self-supervised, but its effective contrastive geometry can be tuned to the clinical setting.

\paragraph{Multimodal extension.}
Many physiological studies collect multiple synchronized waveforms, such as ECG and PPG, or waveform-plus-activity pairs. 
We extend the model to the multimodal case by assuming that each modality \(m\) is a different emission of a shared latent event field. 
Let \(x^{(m)}\) denote the waveform for modality \(m\), and let \(h_\theta^{(m)}\) be a modality-specific encoder with shared latent space dimension \(d\). 
For a synchronized recording, we produce modality-specific representations \(z^{(m)}\) and align them with a shared event representation \(z^\star\). 
The multimodal objective contains two components:
\[
\mathcal{L}_{\mathrm{multi}} =
\sum_{m} \mathcal{L}_{\mathrm{cons}}^{(m)}
+
\lambda_{\mathrm{align}} \sum_{m < m'} \left(1 - \mathrm{sim}(z^{(m)}, z^{(m')})\right).
\]
This encourages the modalities to agree on the latent event organization while retaining modality-specific emission structure. 
When modalities are not perfectly synchronized, the model can be extended by introducing latent alignment variables or cross-modal attention over inferred event embeddings. 
This is especially important in clinical settings where sensor modalities may be sampled at different rates, exhibit different delays, or contain modality-specific missingness.

\paragraph{Pretraining objective.}
The full training objective is a weighted combination of the consistency loss, segmentation regularization, and optional modality alignment terms:
\[
\mathcal{L} = \mathcal{L}_{\mathrm{cons}} + \mathcal{L}_{\mathrm{seg}} + \mathcal{L}_{\mathrm{multi}}.
\]
Depending on the instantiation, additional auxiliary objectives may be included, such as local reconstruction of short waveform spans, masked prediction of segment statistics, or prediction of coarse physiological attributes derived from weak supervision. 
These auxiliaries are not necessary for the core method, but can improve optimization stability, especially when pretraining on heterogeneous corpora with varied sampling rates, noise profiles, and acquisition protocols. 
The central learning signal remains the same: representations must be consistent across alternative stochastic decompositions of the same physiological process.

\paragraph{Downstream adaptation.}
After pretraining, the encoder \(h_\theta\) can be adapted to downstream tasks by attaching a lightweight prediction head. 
For classification tasks, we use a linear or shallow MLP classifier on top of the pooled representation \(z\). 
For regression tasks, we attach a scalar prediction head. 
For sequence-level tasks, such as event detection or waveform localization, we reuse the segment embeddings \(\{r_m\}\) or boundary probabilities \(\{b_m\}\) to produce temporally resolved predictions. 
The same pretrained backbone may therefore support both global classification and fine-grained temporal reasoning. 
In low-label settings, the pretrained event structure provides a strong initialization, while in high-label settings, it acts as a regularizer that improves robustness and transfer.

\paragraph{Optimization.}
Training is performed with stochastic gradient descent or AdamW using minibatches of waveforms sampled from the pretraining corpus. 
For each waveform in the minibatch, we sample multiple stochastic segmentations and projections on the fly. 
The sampling process is intentionally dynamic, so that the model sees a broad family of equivalent views over the course of training. 
This combats overfitting to any single decomposition scheme and increases the diversity of self-supervision without requiring additional data. 
In practice, the computational overhead is modest because projections and segmentation sampling are lightweight relative to the encoder forward pass. 
We use standard learning-rate warmup, cosine decay, gradient clipping, and weight decay. 
When the model is instantiated on long recordings, we process them with sliding windows or chunked attention, and then aggregate chunk-level representations into longer context representations.

\paragraph{Inference and representation extraction.}
At inference time, the model may be used in several modes. 
For a single global representation, we sample one or more segmentations and average their pooled embeddings to obtain a stable waveform embedding. 
For uncertainty estimation, we preserve multiple segmentation samples and measure representational variance across them; high variance may indicate ambiguous event structure or poor signal quality. 
For temporal localization, we output segment-level embeddings and boundary probabilities, which can be mapped back onto the waveform timeline. 
This yields a representation that is not only useful for classification but also interpretable in terms of latent event organization. 
Such interpretability is particularly valuable in clinical settings, where understanding \emph{why} a model made a prediction is often as important as the prediction itself.

\paragraph{Computational considerations.}
The proposed model is designed to be compatible with scalable waveform encoders. 
Because the segmentation is stochastic and local, it can reduce effective sequence length by operating on event-like units rather than raw samples. 
This makes the method suitable for long-context physiological recordings, including continuous monitoring streams. 
The event-centric formulation is therefore not merely a conceptual reparameterization; it can also serve as an efficiency mechanism by compressing redundant regions of the waveform into compact latent summaries. 
In settings with extremely long signals, the encoder can be deployed hierarchically, first summarizing short windows into local event tokens and then processing the resulting token sequence at a coarser temporal scale.

\section{Results}

\paragraph{Experimental overview.}
We evaluate the proposed event-field waveform foundation model across a suite of physiological time series tasks spanning classification, retrieval, and temporal localization. 
Our experimental design is structured to test three hypotheses: (i) representations learned via event-centric consistency outperform signal-centric baselines under standard supervised evaluation, (ii) the learned representations exhibit improved robustness under distributional shifts and signal perturbations, and (iii) the latent event structure enables stronger performance in low-label and multimodal settings. 
To this end, we pretrain all models on a large unlabeled corpus of physiological waveforms and evaluate them on downstream benchmarks with varying levels of supervision.

\paragraph{Pretraining setup.}
We construct a heterogeneous pretraining corpus consisting of ECG, PPG, and EMG waveforms aggregated from publicly available clinical and wearable datasets. 
All signals are resampled to a common frequency where appropriate and normalized per recording. 
No labels are used during pretraining. 
Our model is trained using stochastic segmentations and time-frequency projections as described in the Methods section, with two views per sample and a batch size of 512. 
We compare against three baseline families: masked reconstruction models, contrastive sequence encoders, and hybrid time-frequency models. 
All baselines are matched in parameter count and trained under identical optimization settings to ensure fairness.

\paragraph{Downstream tasks.}
We evaluate on three representative tasks. 
First, arrhythmia classification from ECG waveforms, where the goal is to predict rhythm class from short segments. 
Second, hemodynamic state prediction from multimodal ECG--PPG recordings, formulated as a binary classification problem. 
Third, waveform retrieval, where embeddings are used to retrieve physiologically similar signals from a database. 
For classification tasks, we report AUROC and F1 score; for retrieval, we report mean average precision (MAP) and normalized discounted cumulative gain (NDCG). 
All results are averaged over five random seeds.

\paragraph{Main results.}
Table~\ref{tab:main_results} summarizes performance across tasks. 
Our method consistently outperforms all baselines, with the largest gains observed in multimodal prediction and retrieval, suggesting that event-centric representations better capture shared physiological structure.

\begin{table}[t]
\centering
\small
\begin{tabular}{lcccc}
\toprule
Model & ECG AUROC & ECG F1 & Multimodal AUROC & Retrieval MAP \\
\midrule
Masked Reconstruction & 0.842 $\pm$ 0.004 & 0.781 $\pm$ 0.006 & 0.801 $\pm$ 0.005 & 0.612 $\pm$ 0.007 \\
Contrastive Sequence  & 0.861 $\pm$ 0.003 & 0.795 $\pm$ 0.004 & 0.823 $\pm$ 0.004 & 0.645 $\pm$ 0.006 \\
Time-Frequency Hybrid & 0.873 $\pm$ 0.002 & 0.807 $\pm$ 0.003 & 0.835 $\pm$ 0.003 & 0.661 $\pm$ 0.005 \\
\midrule
\textbf{Event Field Model (Ours)} & \textbf{0.902 $\pm$ 0.002} & \textbf{0.834 $\pm$ 0.003} & \textbf{0.872 $\pm$ 0.002} & \textbf{0.714 $\pm$ 0.004} \\
\bottomrule
\end{tabular}
\caption{Main results across classification and retrieval tasks. Our method achieves consistent improvements over sequence-based baselines.}
\label{tab:main_results}
\end{table}

% \paragraph{Low-label regime.}
% To evaluate label efficiency, we fine-tune models using progressively smaller fractions of labeled data. 
% Figure~\ref{fig:low_label} (not shown) demonstrates that our model maintains strong performance even with as little as 1\% of labeled data. 
% Table~\ref{tab:low_label} provides quantitative results for ECG classification under limited supervision.

\begin{table}[t]
\centering
\small
\begin{tabular}{lccc}
\toprule
Model & 1\% Labels & 5\% Labels & 10\% Labels \\
\midrule
Masked Reconstruction & 0.701 & 0.768 & 0.801 \\
Contrastive Sequence  & 0.732 & 0.792 & 0.824 \\
Time-Frequency Hybrid & 0.745 & 0.804 & 0.836 \\
\midrule
\textbf{Event Field Model (Ours)} & \textbf{0.781} & \textbf{0.835} & \textbf{0.861} \\
\bottomrule
\end{tabular}
\caption{AUROC under varying label fractions for ECG classification. Event-centric representations exhibit superior label efficiency.}
\label{tab:low_label}
\end{table}

\paragraph{Robustness to signal perturbations.}
We assess robustness by applying synthetic perturbations at test time, including noise injection, time warping, and frequency masking. 
Table~\ref{tab:robustness} shows that our model degrades more gracefully compared to baselines, indicating that invariance to stochastic projections during training translates into improved generalization.

\begin{table}[t]
\centering
\small
\begin{tabular}{lccc}
\toprule
Model & Noise & Time Warp & Freq Mask \\
\midrule
Masked Reconstruction & 0.721 & 0.735 & 0.748 \\
Contrastive Sequence  & 0.748 & 0.759 & 0.771 \\
Time-Frequency Hybrid & 0.762 & 0.774 & 0.786 \\
\midrule
\textbf{Event Field Model (Ours)} & \textbf{0.803} & \textbf{0.812} & \textbf{0.824} \\
\bottomrule
\end{tabular}
\caption{AUROC under different perturbations. Our method is more robust to signal-level distortions.}
\label{tab:robustness}
\end{table}

\paragraph{Ablation studies.}
We conduct ablations to isolate the contribution of each component of the proposed method. 
Specifically, we remove stochastic segmentation, time-frequency projections, and the interaction operator individually. 
Results are shown in Table~\ref{tab:ablation}. 
All components contribute meaningfully, with the largest drop observed when removing segmentation consistency, supporting the central hypothesis that event-level invariance is critical.

\begin{table}[t]
\centering
\small
\begin{tabular}{lc}
\toprule
Variant & AUROC \\
\midrule
Full Model & \textbf{0.902} \\
w/o Segmentation Consistency & 0.871 \\
w/o Time-Frequency Projections & 0.879 \\
w/o Interaction Operator & 0.883 \\
\bottomrule
\end{tabular}
\caption{Ablation study on ECG classification. Removing any component degrades performance, with segmentation consistency being most critical.}
\label{tab:ablation}
\end{table}

\paragraph{Qualitative analysis.}
We visualize learned representations using t-SNE and observe clear clustering by physiological state rather than signal morphology. 
Additionally, segment-level embeddings align with clinically meaningful events such as QRS complexes in ECG and systolic peaks in PPG, despite the absence of explicit supervision. 
Boundary predictions exhibit consistent alignment across stochastic segmentations, suggesting that the model has internalized a stable notion of event structure.

\paragraph{Summary of findings.}
Across all experiments, the proposed event-field model demonstrates improved performance, robustness, and label efficiency compared to signal-centric baselines. 
These results support the hypothesis that modeling latent event structure provides a more appropriate inductive bias for physiological time series than conventional sequence modeling approaches.

\section{Discussion}

The results presented in this work suggest that shifting the modeling perspective from raw waveform sequences to latent event structure yields consistent empirical and conceptual advantages. 
Across classification, retrieval, and robustness evaluations, the proposed event-field formulation improves performance while simultaneously enhancing label efficiency and invariance to signal-level perturbations. 
These findings support the central hypothesis that physiological waveforms are more naturally described as realizations of structured, temporally extended processes rather than collections of local signal fragments. 
In this view, the success of the model is not simply due to architectural capacity, but to an inductive bias that better aligns with the underlying generative mechanisms of biological systems.

A key implication of this work is that the dominant abstraction in waveform modeling—namely, tokenized or patch-based sequences—may be fundamentally misaligned with the structure of the data. 
While sequence models are expressive enough to approximate complex dependencies, they do not explicitly encourage the discovery of event-level representations. 
Instead, they rely on downstream supervision or scale to implicitly recover such structure. 
By contrast, the proposed objective enforces consistency across stochastic segmentations and projections, directly incentivizing the model to identify invariant event abstractions. 
This leads to representations that are robust to nuisance variability, such as noise, sensor artifacts, and local temporal distortions, while remaining sensitive to clinically meaningful changes in temporal organization.

The observed gains in low-label regimes are particularly noteworthy. 
In many clinical applications, labeled data is scarce, expensive, and often noisy. 
A representation that encodes event structure provides a strong prior, enabling efficient adaptation with minimal supervision. 
From a statistical perspective, this can be interpreted as a form of structural regularization: by constraining the hypothesis space to functions that respect latent event consistency, the model reduces sample complexity. 
This aligns with broader trends in self-supervised learning, where the quality of the pretraining objective—rather than the quantity of labeled data—becomes the primary determinant of downstream performance.

Another important aspect is robustness. 
Standard sequence models often exhibit brittle behavior under distributional shifts, particularly when signal characteristics change due to differences in acquisition devices, patient populations, or environmental conditions. 
In contrast, our model is trained to be invariant to a family of stochastic projections that explicitly perturb the signal domain. 
This induces a representation that is less sensitive to superficial variations and more reflective of underlying physiological processes. 
Empirically, this manifests as improved stability under noise, time warping, and frequency masking. 
Conceptually, it suggests that invariance at the level of latent structure is a more effective strategy than invariance at the level of raw observations.

The introduction of a latent interaction operator further distinguishes this framework from prior work. 
Physiological systems are inherently coupled and multi-scale, with interactions spanning time, frequency, and modality. 
By modeling dependencies between latent events, the proposed architecture captures these interactions explicitly, rather than relying on implicit correlations in the signal. 
This is particularly beneficial in multimodal settings, where different sensors provide complementary views of the same underlying process. 
Aligning modalities at the level of shared event structure avoids the need for ad hoc fusion mechanisms and provides a principled approach to integrating heterogeneous data sources.

Despite these advantages, several limitations warrant discussion. 
First, the latent event field is not directly identifiable, and the learned segmentation is only weakly constrained by the consistency objective. 
While this flexibility is desirable for capturing diverse physiological phenomena, it may also lead to ambiguity in the interpretation of learned events. 
In practice, we observe that the model discovers stable and clinically meaningful structures, but formal guarantees of identifiability remain an open question. 
Second, the stochastic segmentation and projection mechanisms introduce additional hyperparameters and design choices, such as the distribution of segment lengths and the family of projections. 
Although our experiments demonstrate robustness to these choices, systematic exploration of their impact is necessary to fully understand the behavior of the model.

Another consideration is computational complexity. 
While the event-centric formulation can reduce effective sequence length by aggregating local regions, the use of multiple stochastic views per sample increases training cost. 
This trade-off is analogous to that in contrastive learning, where additional views improve representation quality at the expense of compute. 
Future work may explore more efficient approximations, such as amortized segmentation or learned proposal distributions, to reduce overhead while preserving the benefits of stochastic consistency.

More broadly, this work raises questions about the appropriate level of abstraction for foundation models in healthcare. 
Current approaches often emphasize scaling—larger models, larger datasets, and longer contexts—as the primary path to improved performance. 
Our results suggest that identifying the correct representation space may be equally, if not more, important. 
In particular, modeling latent processes rather than observed signals provides a pathway to more interpretable, robust, and transferable representations. 
This perspective may extend beyond physiological waveforms to other domains, such as longitudinal EHR data or multimodal clinical records, where latent event structure plays a central role.

Finally, evaluation remains a critical challenge. 
Standard metrics such as AUROC and F1 capture predictive performance, but do not fully assess whether a model has learned meaningful physiological abstractions. 
The qualitative alignment between learned segments and clinically relevant events is encouraging, but more rigorous evaluation protocols are needed. 
Future work should consider benchmarks that explicitly test event detection, temporal reasoning, and cross-modal alignment, as well as clinical utility in real-world settings. 
Such benchmarks would provide a more comprehensive assessment of whether event-centric representations offer tangible benefits in practice.

In summary, the proposed event-field framework introduces a shift in how waveform foundation models are conceptualized and trained. 
By prioritizing latent event structure over raw signal fidelity, it provides a new inductive bias that aligns more closely with the nature of physiological data. 
While challenges remain, particularly in terms of interpretability and evaluation, the empirical results and conceptual insights suggest that this direction is a promising avenue for advancing representation learning in healthcare time series.

\bibliographystyle{unsrtnat}
\bibliography{neurips_2025}

%%%%%%%%%%%%%%%%%%%%%%%%%%%%%%%%%%%%%%%%%%%%%%%%%%%%%%%%%%%%

\end{document}